\documentclass[conference]{IEEEtran}
\IEEEoverridecommandlockouts
\usepackage{amsmath,amssymb,amsfonts}
\usepackage{subfigure}
\usepackage[numbers]{natbib}
\usepackage{makecell}  
\usepackage{amsmath} 
\usepackage{amsthm}  
\usepackage{booktabs} 
\usepackage{url}
\newtheorem{definition}{Definition} 
\usepackage{natbib} 
\usepackage{multirow}
\usepackage{algorithm}
\usepackage{algpseudocode}
\usepackage{graphicx}
\usepackage{textcomp}
\usepackage{hyperref}
\usepackage{xcolor}
\def\BibTeX{{\rm B\kern-.05em{\sc i\kern-.025em b}\kern-.08em
    T\kern-.1667em\lower.7ex\hbox{E}\kern-.125emX}}
\begin{document}

\title{TSFeatLIME: An Online User Study in Enhancing Explainability in Univariate Time Series Forecasting\\
\thanks{This work was partially funded by  EPSRC CHAI project (EP/T026707/1) and EPSRC IAA fund at the University of Bristol.}
}


\author{
    \IEEEauthorblockN{Hongnan Ma\textsuperscript{1}, Kevin McAreavey\textsuperscript{2}, Weiru Liu\textsuperscript{2}}
    \IEEEauthorblockA{\textsuperscript{1}School of Computer Science, University of Bristol, Bristol, UK\\
    \textsuperscript{2}School of Engineering Mathematics and Technology, University of Bristol, Bristol, UK\\
    \{ex20249, kevin.mcareavey, weiru.liu\}@bristol.ac.uk}
}

\maketitle

\begin{abstract}
Time series forecasting, while vital in various applications, often employs complex models that are difficult for humans to understand. Effective explainable AI techniques are crucial to bridging the gap between model predictions and user understanding. This paper presents a framework - TSFeatLIME, extending TSLIME, tailored specifically for explaining univariate time series forecasting. TSFeatLIME integrates an auxiliary feature into the surrogate model and considers the pairwise Euclidean distances between the queried time series and the generated samples to improve the fidelity of the surrogate models. However, the usefulness of such explanations for human beings remains an open question. We address this by conducting a user study with 160 participants through two interactive interfaces, aiming to measure how individuals from different backgrounds can simulate or predict model output changes in the treatment group and control group. Our results show that the surrogate model under the TSFeatLIME framework is able to better simulate the behaviour of the black-box considering distance, without sacrificing accuracy. In addition, the user study suggests that the explanations were significantly more effective for participants without a computer science background.
\end{abstract}


\section{Introduction}
Time series data, which consists of sequential records collected over time, plays a crucial role in high-stakes decision-making across various fields such as healthcare \cite{kaushik2020ai} and finance \cite{shi2024counterfactual}. The primary objective of time series forecasting is to predict future observations within a temporally ordered sequence, which may exhibit underlying trends and seasonality patterns. 

Within the realm of time series forecasting, several established models exist, broadly categorized into statistical models (e.g., SARIMA, Exponential Smoothing, Prophet, Theta), modern machine learning models (e.g., tree-based ensembles such as XGBoost, CatBoost, LightGBM, Gaussian Process Regression), including deep learning models (e.g., DeepAR, N-BEATS, Informer, and SCINet). Although these black-box models have shown promise in delivering accurate predictions, they often sacrifice transparency, interpretability, and fairness in their decision-making processes. Consequently, users might find it challenging to understand the reasoning behind a model's predictions, even if the model showcases impressive accuracy. Furthermore, \cite{ribeiro2016should} mentioned that humans as inherently curious beings, have a desire to understand the reasoning underlying the decision-making processes employed by the black-box model. Consequently, individuals are less likely to utilize a machine learning model if they cannot fully understand and trust its operations. Due to these two reasons, researchers delve into the area of explainable AI (XAI).

While substantial progress has been achieved in explainability within the fields of tabular data, computer vision and natural language processing, providing explanations for time series data remains challenging due to its complex non-linear temporal dependencies and the multi-dimensional nature \cite{saluja2021towards}. When dealing with time series data, it is not sufficient to merely identify the importance of each timepoint contributing to a model's decision. Equally important is understanding how features from seasonal patterns, trends, and cyclical behaviours impact the decision-making process of these models.

Theissler et al. \cite{theissler2022explainable} proposed a taxonomy for XAI for time series forecasting, which subdivides the method into time points-based, subsequences-based and instance-based explanations. We primarily concentrate on the time points-based explanation. Within this area, two principal methodologies emerge: attribute-based and attention-based explanations. The attention-based mechanism is a type of ante-hoc explainability method. It is integrated into the architecture of recurrent neural networks, and the insights it provides are immediately accessible after the model's learning phase \cite{rojat2021explainable}. In contrast, for attribute-based methods, there are three subclasses: gradient-based, structure-based, and surrogate-and-sampling-based techniques. The surrogate-and-sampling approach primarily focuses on creating samples adjacent to a specified input to train an interpretable model like LIME (Local Interpretable Model-Agnostic Explanation). Alternatively, SHAP (SHapley Additive exPlanations) employs a game-theoretical framework to assign weights to features for better attribution. For LIME, it is worth noting that LIME has been applied in the subsequence-based domain \cite{schlegel2021ts}, wherein each segment, post-extraction, is treated as a feature. The importance attributed to each of these segments serves as the basis for explanation. Nevertheless, a notable gap remains in applying LIME explanations for time points-based problems. How to provide comprehensive explanations centred on these timepoints continues to be a topic of discussion in the field.

To compare the performance of different local explainers, fidelity, serves as a crucial metric for evaluation in the area of model explainability \cite{guilleme2019agnostic}. Fidelity quantifies the extent to which the explanatory model accurately mirrors the behaviour of the black-box model. 

In this work, we introduce a new framework TSFeatLIME, based on TSLIME, which itself is based on LIME, but tailored for explaining the time series black-box model. TSFeatLIME extracts derived features - auxiliary features related to time points to augment the explainability of the surrogate model. Our experimental findings reveal that incorporating auxiliary features neither significantly increases nor decreases the behaviour of surrogate models when mimicking black-box models. To further enhance the model's fidelity, we also consider the pairwise Euclidean distance between the queried time series and the samples. Introducing this can prioritize proximity samples and disregard those far from the queried time series in the process of perturbation. Our results demonstrate that considering this can dramatically improve the model's ability to mimic without sacrificing black-box accuracy.

Although most studies focus on providing explanations through computational methods, research assessing the quality of explanations based on user evaluations is quite limited. According to the papers surveyed in \cite{nguyen2024human}, only 17\% involve humans in both analysis and evaluation, as required by the human-centered design process. Defining what constitutes an optimal explanation to enable systematic and rigorous evaluation of XAI methods remains an unresolved issue. Given the limited research on evaluating explanations, we conducted a user study on the platform Prolific, engaging 160 participants through two distinct interactive interfaces that we designed: a control group and a treatment group. This design aims to evaluate the effectiveness of our explanation approach and explore whether participants from different backgrounds—computer science and non-computer science—experience significant effects. Furthermore, we also want to measure user satisfaction, trust, or the goodness of an explanation. 
 
To design the exercises that evaluate the TSFeatLIME framework within the interface, we adopt the concept of \textit{simulatability} from \cite{doshi2017towards} which is regarded as a proxy for evaluating explainability. Specifically, a model is deemed simulatable if an individual can predict the model's behaviour on new inputs \cite{doshi2017towards}. Counterfactual simulation is a type of simulation which is applied in this scenario. The details will be further discussed in the User Study section. The contributions of this paper are as follows:
\begin{itemize}
\item[$\bullet$] We proposed TSFeatLIME for training a surrogate model to explain time series forecasting results. By extracting auxiliary features such as rolling and expanding windows from the perturbed data, we provide comprehensive explanations for users. Additionally, considering the distance between the queried and sample time series increases the fidelity of the local surrogate model.
\item[$\bullet$] We developed an online web-based study, integrating with Prolific, to assess whether the provided explanations serve as a useful form of XAI for end-users.
\end{itemize}. 

\section{Related work}
\label{relatedwork}

\begin{figure*}[htbp]
\centering
\subfigure[LIME]{
\begin{minipage}[t]{0.50\linewidth}
\label{lime}
\centering
\includegraphics[width=0.9\linewidth]{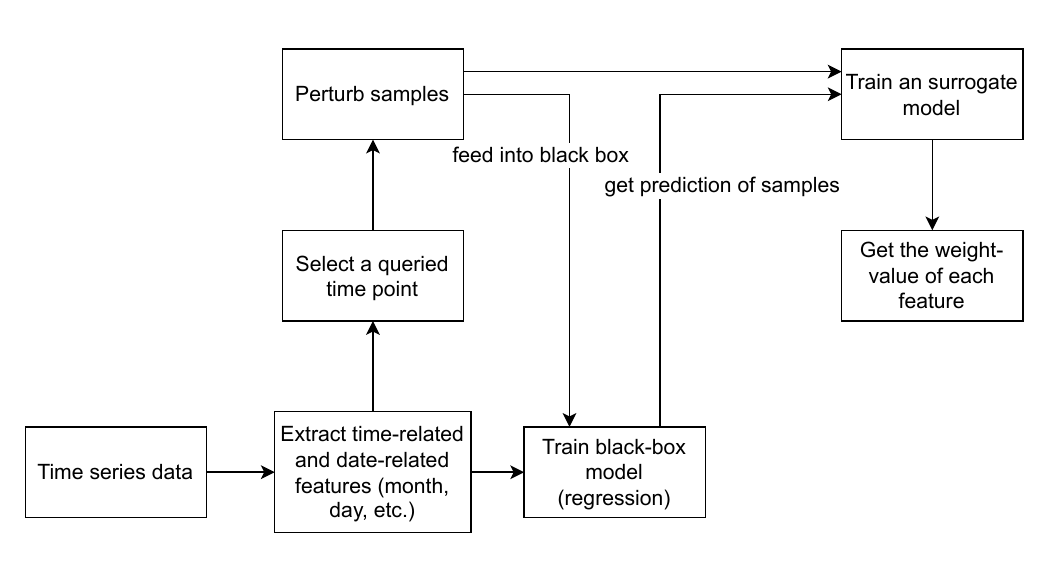}
\end{minipage}%
}%
\subfigure[TSLIME]{
\begin{minipage}[t]{0.40\linewidth}
\label{tslime}
\centering
\includegraphics[width=1.0\linewidth]{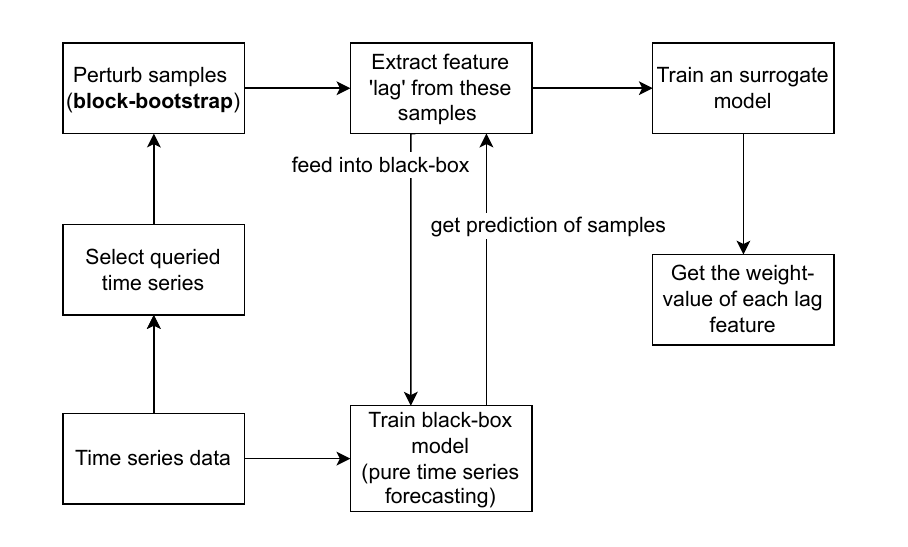}
\end{minipage}%
}%
\caption{LIME and TSLIME for time series}
\end{figure*}

XAI is still a young field in the area of time series, with most works tilting towards classification problems \cite{theissler2022explainable}. Bridging the gap in explainability for time series forecasting, several approaches have been explored. The intersection of time points-based explanation and LIME of explainability method has seen trends where time series data undergoes transformation into a feature-based format, thereby shifting the problem from a pure forecasting scenario to a traditional regression problem and feeding it into black-box models. For example, a study by \cite{saluja2021towards} introduces the lag features and leverages Support Vector Regression (SVR) for model training. Similarly, \cite{narang2023interpretable} and \cite{ma2022explainable} extract both date-related and time-related features from original time series data and then feed them into the model-building phase. Figure \ref{lime} demonstrates the process of LIME applied in time series. By extracting these features, the goal is often to simplify the process of generating explanations, allowing for the direct application of LIME. However, these approaches have their limitations. Utilizing the regression model directly makes the black-box model less accurate because the regression model does not capture the dependencies between timepoints. The perturbation phase in LIME isn't specifically tailored to time points-based. After the feature extraction, the data undergoes perturbation which is similar to traditional tabular data. It can blur the temporal dependencies inherent in raw time series data when generating new samples to train a local model.

In the study by \cite{aix360-sept-2019}, the authors introduced the Time Series Local Interpretable Model-agnostic Explainer (TSLIME), tailored specifically for time points-based explanations. TSLIME is an extension of LIME focusing solely on pure forecasting methods without requiring additional feature extraction before training the black-box model. Figure \ref{tslime} demonstrates the process of TSLIME explanation. A notable distinction of TSLIME from prior work is its use of the time series perturbation method, \textit{block bootstrap}, to generate local input perturbation samples. Furthermore, it then utilizes a multilinear surrogate model for explanations, optimally approximating the model's response.

While TSLIME represents a significant advancement compared to previous methods, its approach of using only lag features to construct the local explainer might lead to explanations that could be unclear to users. Furthermore, ignoring the distance between the queried time series and generated samples in the perturbation process also affects the fidelity of the surrogate model. Our objective is to identify and incorporate a wider range of time features, enhancing the depth of information provided to non-experts.

\section{TSFeatLIME}
\label{tsfeatlime}
\subsection{Perturbation for time series}
Perturbation, aimed to generate supplementary samples, is prevalent both in machine learning and local explanation domains. In LIME, perturbation is employed to produce a localized dataset surrounding the queried data point. For tabular data, LIME generates new samples by individually perturbing each feature, drawing values from a normal distribution defined by the feature's mean and standard deviation. However, time series data, with its unique structure, necessitates a different approach: \textit{block bootstrap}. The bootstrap's fundamental idea, introduced by \cite{efron1992bootstrap}, is to estimate desired metrics by resampling—with replacement—from the original dataset. However, applying the bootstrap to time series data presents challenges. This is primarily because time series data is presumed to exhibit autocorrelation, where noise is correlated between adjacent data points. Randomly shuffling this data disrupts these inherent correlations. To address the limitations of bootstrapping with time series data, the block bootstrap method \cite{kreiss2012bootstrap} extends Efron's work \cite{efron1992bootstrap}, which is initially tailored for traditional independent and identically distributed (i.i.d) data. Instead of resampling individual observations, the block bootstrap resamples chunks or `blocks' of continuous observations. In our framework, we employ the Moving Block Bootstrap (MBB) — a variant of the block bootstrap. 

\begin{figure*}[htbp]
\centering
\includegraphics[scale=0.37]{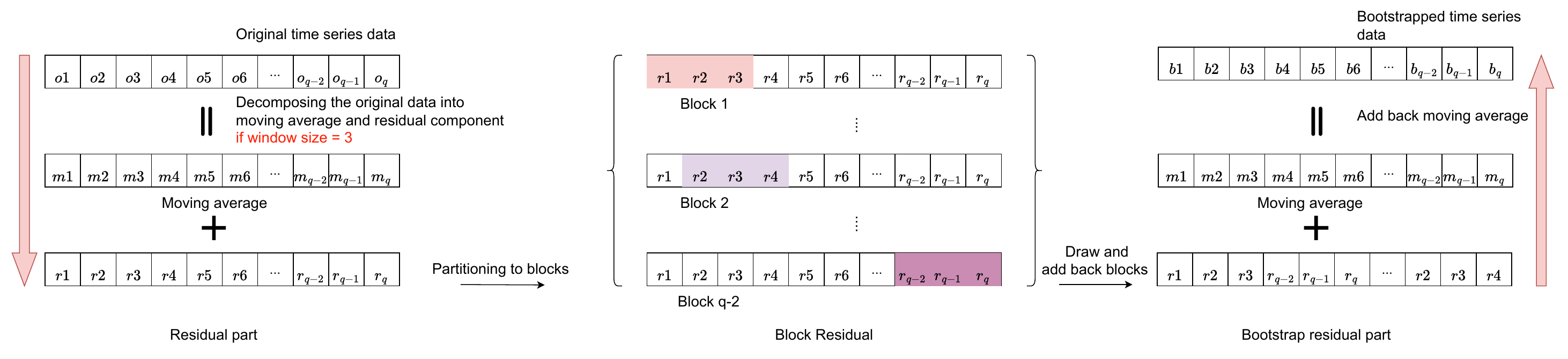}
\caption{Illustration of the process of moving block bootstrapping (MBB)} 
\label{bootstrap}
\end{figure*}

Figure \ref{bootstrap} illustrates the entire process of applying the MBB to time series data to generate more samples. Specifically, the first step of MBB is to decompose the original temporal time series data $O$ into moving average $M$ and residual components $R$. These two components have the same dimension with $O$. The Moving average $M$ is a statistical calculation derived from $O$ that serves to smooth variations in the dataset. This property reduces fluctuations, thus providing a clearer view of the underlying trend. The Residual component $R$, on the other hand, represents the differences remaining after the moving average has been subtracted from $O$. This component effectively captures variations or shifts around the established trend. Subsequent steps involve grouping the residuals from the prior phase into a contiguous series. Using a sliding window approach, as depicted in the left and center sections of Figure \ref{bootstrap}, and given that the length of $O$ is $q$ and the \textbf{block length} is $l$, we derive $q-l+1$ sets of continuous residuals. Perturbations are applied to residuals after removing the moving average, rather than on $O$. This method's advantage is that by perturbing only the residual component, the algorithm introduces localized changes that don't drastically alter the overall behaviour of time series. After that, the process involves randomly selecting two blocks and exchanging them to obtain the bootstrap residual component. The number of swap blocks is determined by the \textbf{block swap}. Subsequently, $M$ is added back to generate a time series data sample $B_i$. This `draw and add back' process is repeated $p$ times to generate sample set $B=\{B_1, B_2, \dots, B_p\}$. Once it is done with these repetitions, the block bootstrap-based perturbation is complete. In this example, $O = \{o_1,o_2,\dots,o_q\}$ represents the queried time series, while $B_i = \{b_{i1}, b_{i2}, \ldots, b_{iq}\}$ represents the sample that has been perturbed. The $O$ and $B_i$ each have a length of $q$.

\subsection{Auxiliary features}
We introduced a set of interpretable auxiliary features to enrich the surrogate model's ability to describe the underlying time dynamics. The set of features could be defined by the expert user as well based on their demand. Apart from lag, we also define two features: Rolling window, and Extending window. 

\begin{definition}
The lag feature with offset $k \in \mathbb{N}$ is defined for each $t = 1, 2, \dots, n$ as:
\end{definition}

\begin{small}
\begin{equation*}
   \text{Lag}_{k}(t) =
	\begin{cases}
		y_{t-k} & \text{if } k < t \\
		\text{undefined} & \text{otherwise}
	\end{cases}
\end{equation*}
\end{small}

Lagged features introduce historical values from previous timepoints as new features. $n$ represents the total number of observations in a time series. $y_t$ represents the value at time point $t$.

\begin{definition}
The rolling window feature with offset $k \in \mathbb{N}$ and window length $w \in \mathbb{N}$ is defined for each $t = 1, 2, \dots, n$ as:
\end{definition}

\begin{small}
\begin{equation*}
	\text{RollingWindow}_{k, w}(t) = 
	\begin{cases}
		\frac{1}{w} \sum_{i = t-k-w+1}^{t-k} y_{i} & \text{if valid}(k,w,t) \\
		\text{undefined} & \text{otherwise}
	\end{cases}
\end{equation*}
\end{small}
where $\text{if valid}(k,w,t) = k < t  \text{ and } w \le t-k.$

Given a fixed window size $w$, rolling window features refer to a time period prior to time point $t$ spanning over $w$ time points. 

\begin{definition}
The expanding window feature with window length $w \in \mathbb{N}$ is defined for each $t = 1, 2, \dots, n$ as:
\end{definition}

\begin{small}
\begin{equation*}
	\text{ExpandingWindow}_{w}(t) =
	\begin{cases}
		\frac{1}{w} \sum_{i = t-w}^{t-1} y_{i} & \text{if } w < t \\
		\text{undefined} & \text{otherwise}
	\end{cases}
\end{equation*}
\end{small}

Expanding window features compute statistics over available observations up to the current time point, effectively `expanding' as time progresses. It is similar to the rolling window, but its start point is fixed at the $t-1$.

Suppose there is a toy time series dataset with six time-based observations (n=6): 10, 20, 30, 40, 50 and 60. The feature and auxiliary feature values can be seen in Table \ref{featureList}. 

\begin{table*}[ht]
\centering
\scriptsize 
\caption{Toy example for feature and auxiliary features}
\label{featureList}
\begin{tabular}{
  >{\raggedright\arraybackslash}p{0.2cm} 
  >{\raggedright\arraybackslash}p{0.2cm} 
  >{\raggedright\arraybackslash}p{0.7cm} 
  >{\raggedright\arraybackslash}p{0.7cm} 
  >{\raggedright\arraybackslash}p{0.7cm} 
  >{\raggedright\arraybackslash}p{0.7cm} 
  >{\raggedright\arraybackslash}p{0.7cm} 
  >{\raggedright\arraybackslash}p{0.7cm} 
  >{\raggedright\arraybackslash}p{0.7cm} 
  >{\raggedright\arraybackslash}p{0.7cm} 
  >{\raggedright\arraybackslash}p{0.75cm} 
  >{\raggedright\arraybackslash}p{0.75cm} 
  >{\raggedright\arraybackslash}p{0.75cm} 
  >{\raggedright\arraybackslash}p{0.75cm} 
  >{\raggedright\arraybackslash}p{0.75cm} 
} 
\toprule
 &  & \multicolumn{5}{c}{$\text{Lag}_{k}(t)$} & \multicolumn{3}{c}{$\text{RollingWindow}_{k, w}(t) [w=3]$} & \multicolumn{4}{c}{$\text{ExpandingWindow}_{w}(t)$} \\
\cmidrule(lr){3-7} \cmidrule(lr){8-10} \cmidrule(lr){11-15}
$t$   &  $y_{t}$ & $k=1$ & $k=2$ & $k=3$ & $k=4$ & $k=5$ & $k=1$ & $k=2$ & $k=3$ & $w=1$ & $w=2$ & $w=3$ & $w=4$ & $w=5$ \\
\midrule
 1 & 10 & - &  - & - & - & - & - & - & - & - & - & - & - & -  \\
 2 & 20 & 10 &  - & - & - & - & - & - & - & 10 & - & - & - & - \\
 3 & 30 & 20 & 10 & - & - & - & - & - & - & 20 & 15 & - & - & -  \\
 4 & 40 & 30 & 20 & 10 & - & - & 20 & - & - & 30 & 25 & 20 & - & - \\
 5 & 50 & 40 & 30 & 20 & 10 & - & 30 & 20 & - & 40 & 35 & 30 & 25 & - \\
 6 & 60 & 50 & 40 & 30 & 20 & 10 & 40 & 30 & 20 & 50 & 45 & 40 & 35 & 30 \\
\bottomrule
\end{tabular}
\end{table*}

\subsection{TSFeatLIME explanation tool}
The MBB perturbation process is discussed in Figure \ref{bootstrap}, and hence, the details of generating the perturbed sample set $B$ by applying MBB function to the queried time series $O$ are omitted in the Pseudocode \ref{alg:algorithm}. The perturbed sample set $B$ consists of $p$ samples. That is $B=\{B_1, B_2, \dots, B_p\}$. Each sample $B_i$ and the queried time series $O$ consist of $q$ timepoints. That is, $B_i = \{b_{i1}, b_{i2}, \dots, b_{iq}\}$ and $O = \{o_1,o_2,\dots,o_q\}$. Here, $b_{ij}$ is defined as the $j^{\text{th}}$ time point in sample $B_i$. The Euclidean distance between $O$ and $B_i$ is represented as:

\begin{small}
\begin{equation}
d(O, B_i) = \sqrt{\sum_{j=1}^q (o_{j} - b_{ij})^2}
\end{equation}
\end{small}

The distance $d$ is assigned to each sample $B_i$ in $B$ based on their proximity to the queried time series being explained, such that samples closer to the time point receive higher values, thereby ensuring the explanations are more locally faithful and relevant. The pairwise Euclidean distances set $D$ is defined in Line 2, which contains all distances between the queried time series and generated samples. Each generated sample $B_i$ is fed into the black-box model $f$ to obtain the predicted value $f{(B_i)}$. Then put each $f{(B_i)}$ in list $y_{perturb}$ for training the surrogate model in Line 3. Subsequently, the feature or auxiliary feature function $a$ is applied on each sample $B_i$ to obtain a series of feature values, represented as $a(B_{i(q+1)})$. Details about how to generate feature or auxiliary feature values using $a$ can be found in the previous subsection. The same process is applied to all generated samples; we combine all feature values for each sample and aggregate them into $a_{set}$ in Line 4. Finally, the surrogate model $g$ is trained by feeding it ${a_{set}}$, $y_{perturb}$, and $D$ in Line 5 and 6. The weights of the feature or auxiliary feature are extracted from the coefficient of $g$, where $L$ is a vector recording these weights. 
\begin{algorithm}[tb]
\caption{Pseudocode of TSFeatLIME}
\label{alg:algorithm}
\textbf{Input}: Queried time series $O = \{o_1, o_2, \dots, o_q\}$, Black-box model $f$, (Auxiliary) feature function $a$\\
\textbf{Output}: Surrogate model $g$, (Auxiliary) feature weights $L$
\begin{algorithmic}[1]
\State $B \gets \Call{MBB}{O}$
\State $D \gets [d(O, B_1), d(O, B_2), \ldots, d(O, B_p)]$
\State $y_{perturb} \gets [f(B_1), f(B_2), \ldots, f(B_p)]$
\State $a_{set} \gets [a(B_{1(q+1)}), a(B_{2(q+1)}), \ldots, a(B_{p(q+1)})]$
\State $g, L \gets \Call{LinearRegression}{a_{set}, y_{perturb}, D}$
\State \textbf{Return} $g$, $L$
\end{algorithmic}
\end{algorithm}

\section{Experiment of setup and Evaluation}
\label{experimentofevaluation}
\subsection{Datasets and forecasting models}
We validated TSFeatLIME using two univariate time series forecasting datasets. The Superstore sales dataset from Community.tableau.com (2017) records sales from 2014 to 2017, featuring nearly 10,000 data points across 21 distinct features. We focus on furniture sales. The second dataset comprises daily records of electricity demand in Spain from 2014 to 2018. Both of them are resampled on a monthly frequency, employing average daily consumption values and setting the start of each month as the index. \cite{ensafi2022time} compared forecasting models such as SARIMA, Triple Exponential Smoothing, Prophet, Vanilla LSTM, and CNN. We evaluated these models on our datasets using Root Mean Square Error (RMSE) and Mean Absolute Percentage Error (MAPE), where lower values indicate higher accuracy. The stacked LSTM emerged as the best model for furniture sales forecasting, while the Vanilla LSTM excelled in electricity consumption forecasting. Consequently, both models are used as black-box models in subsequent experiments. Details about data partitioning and training results are available under GitHub: \url{https://github.com/ts-xai/whyforecast}.
\subsection{Fidelity metrics}
The fidelity of an explanation on the time series can be represented as:

\begin{small}
\begin{equation}
\text{Fidelity}^{L} (g \mid f, N_{test}) = L(g \mid f, N_{test})
\end{equation}
\end{small}

Here, $L$ donates the performance metric, $g$ represents the surrogate model and $f$ denotes the black-box model. Taking the RMSE as a performance metric, and calculating the fidelity of an explanation on the queried time series $O_i$ across $|N_{test}|$ testing samples can be represented as:

\begin{small}
\begin{equation}
\text{Fidelity}^{RMSE} (g \mid f,N_{test})= \sqrt{\frac{1}{|N_{test}|}\sum_{O_i \in N_{test}}^{}[f(O_i) - g(O_i)]^2}
\end{equation}
\end{small}

In the context of our study, fidelity measures whether the behaviour of the mimic the black box model is enhanced by incorporating additional time-based auxiliary features and considering the distance of perturbation samples in the testing dataset $N_{test}$.

In our experiment, we will explore the impact of varying key parameters, specifically `block length' and `block swap' in the perturbation process - MBB, on the performance of TSFeatLIME. We considered block lengths of 3,4 and 5, and block swaps of 2,3 and 4. Table \ref{hyperparameter} demonstrates the best hyperparameter selection for two datasets. Finally, we selected the hyperparameters for both datasets: a block length of five and a block swap of two. Considering the fixed block length of five, block swap of two and a sample size of 1,000 for training the surrogate model, we implemented an evaluation methodology comprising five distinct iterations to mitigate variance in sample generation and surrogate model training. The fidelity metric $\text{Fidelity}^{MAE}$, $\text{Fidelity}^{MSE}$ and $\text{Fidelity}^{MAPE}$ used suggests that a lower value of $\text{Fidelity}^{L}$ is indicative of superior results. In Table \ref{weight}, horizontal comparison results show that for all datasets and all types of auxiliary surrogate models, adding distance clearly improves the fidelity of the surrogate model. Furthermore, vertical comparison results reveal that the fidelity of the surrogate model across any of the auxiliary features does not vary significantly. This suggests that adding auxiliary features can provide additional information without affecting the surrogate model's fidelity.

\begin{table*}[ht]
\centering
\scriptsize 
\caption{Hyperparameter selection for sales and electricity consumption (Lag, RW, and EW)}
\label{hyperparameter}
\begin{tabular}{@{}lllccc ccc ccc@{}}
\toprule
& \multicolumn{10}{c}{Metrics by Auxiliary Feature} \\
\cmidrule(l){4-12}
Dataset & Block len & Swap & \multicolumn{3}{c}{Lag} & \multicolumn{3}{c}{RW} & \multicolumn{3}{c}{EW} \\ 
\cmidrule(lr){4-6} \cmidrule(lr){7-9} \cmidrule(l){10-12}
& & & MAE & RMSE & MAPE & MAE & RMSE & MAPE & MAE & RMSE & MAPE \\ 
\midrule

& 3 & 2 & 0.017 & 0.032 & 8.08\% & 0.016 & 0.029 & 8.34\% & 0.019 & 0.026 & 9.86\% \\
&   & 3 & 0.026 & 0.059 & 15.46\% & 0.025 & 0.055 & 12.59\% & 0.030 & 0.050 & 12.14\% \\
&   & 4 & 0.035 & 0.080 & 0.034 & 0.073 & 16.04\% & 15.54\% & 0.040 & 0.068 & 19.46\% \\

Sales & 4 & 2 & 0.011 & 0.017 & 4.88\% & 0.010 & 0.015 & 4.89\% & 0.011 & 0.013 & 5.98\% \\
&   & 3 & 0.021 & 0.041 & 8.30\% & 0.020 & 0.039 & 8.25\% & 0.021 & 0.029 & 10.21\% \\
&   & 4 & 0.025 & 0.045 & 10.58\% & 0.025 & 0.046 & 10.52\% & 0.024 & 0.032 & 14.20\% \\

& 5 & 2 & \underline{0.007} & \underline{0.012} & \underline{2.41\%} & \underline{0.006} & \underline{0.010} & \underline{2.32\%} & \underline{0.008} & \underline{0.012} & \underline{3.48\%} \\
&   & 3 & 0.019 & 0.035 & 5.23\% & 0.019 & 0.035 & 5.62\% & 0.017 & 0.025 & 7.94\% \\
&   & 4 & 0.023 & 0.041 & 8.64\% & 0.023 & 0.041 & 9.39\% & 0.023 & 0.034 & 11.24\% \\

& 3 & 2 & 0.006 & 0.007 & 1.27\% & 0.010 & 0.012 & 2.27\% & 0.014 & 0.017 & 3.04\% \\
&   & 3 & 0.009 & 0.011 & 1.95\% & 0.013 & 0.015 & 2.97\% & 0.019 & 0.023 & 4.15\% \\
&   & 4 & 0.011 & 0.014 & 2.38\% & 0.016 & 0.020 & 3.58\% & 0.022 & 0.027 & 4.71\% \\

Electricity & 4 & 2 & 0.005 & 0.007 & 1.22\% & 0.008 & 0.009 & 1.73\% & 0.010 & 0.012 & 2.17\% \\
&   & 3 & 0.008 & 0.010 & 1.74\% & 0.012 & 0.015 & 2.72\% & 0.016 & 0.020 & 3.48\% \\
&   & 4 & 0.013 & 0.019 & 2.78\% & 0.019 & 0.024 & 4.17\% & 0.026 & 0.036 & 5.38\% \\

& 5 & 2 & \underline{0.004} & \underline{0.006} & \underline{0.96\%} & \underline{0.007} & \underline{0.008} & \underline{1.50\%} & \underline{0.009} & \underline{0.011} & \underline{2.08\%} \\
&   & 3 & 0.008 & 0.011 & 1.76\% & 0.012 & 0.015 & 2.69\% & 0.016 & 0.021 & 3.44\% \\
&   & 4 & 0.008 & 0.012 & 1.80\% & 0.014 & 0.016 & 3.00\% & 0.020 & 0.028 & 4.26\% \\
\bottomrule
\end{tabular}%
\end{table*}

\begin{table*}[]
\centering 
\scriptsize 
\caption{Metric values of different models with and without distance}
\label{weight}
\begin{tabular}{lccccccc}
\toprule
\multirow{2}{*}{Dataset} & \multirow{2}{*}{Method} & \multicolumn{3}{c}{Without distance} & \multicolumn{3}{c}{Distance} \\
\cmidrule(lr){3-5} \cmidrule(lr){6-8}
                         &                         & MAE      & RMSE     & MAPE    & MAE      & RMSE     & MAPE    \\
\midrule
\multirow{3}{*}{Sales}   & TSFeatLIME (Lag)        & 0.0241   & 0.0453   & 8.13\%  & 0.0069   & 0.0115   & \underline{2.41\%}  \\
                         & TSFeatLIME (RW)         & 0.0235   & 0.0395   & 9.02\%  & 0.0064   & 0.0098   & \underline{2.32\%}  \\
                         & TSFeatLIME (EW)         & 0.0342   & 0.0477   & 9.11\%  & 0.0080   & 0.0118   & \underline{3.48\%}  \\
\addlinespace
\multirow{3}{*}{Electricity} & TSFeatLIME (Lag)     & 0.0080   & 0.0113   & 1.77\%  & 0.0039   & 0.0047   & \underline{0.91\%}  \\
                             & TSFeatLIME (RW)      & 0.0112   & 0.0148   & 2.45\%  & 0.0066   & 0.0077   & \underline{1.50\%}  \\
                             & TSFeatLIME (EW)      & 0.0131   & 0.0167   & 2.89\%  & 0.0095   & 0.0113   & \underline{2.08\%}  \\
\bottomrule
\end{tabular}
\end{table*}

\section{User study}
\label{interfacedesign}
\subsection{Methodology}
\subsubsection{Interface design}
To investigate the impact of explanation on user understanding of the process of black-box, we designed a comparative study with two different interfaces which integrate with Prolific. This integration allows us to recruit participants from Prolific based on our specific requirements. The control group accessed an interface without additional explanatory content while the treatment group's interface included both visual and textual explanations. Participants from different groups accessed these interfaces through separate links. After providing consent, participants could continue or exit the experiment. A tutorial was provided afterwards. Following the tutorial, participants completed four exercises. After completing all exercises, participants were presented with a questionnaire. The interface for consent, the tutorial, the questionnaire, and the source code are also available in our \href{https://github.com/yourusername/yourrepository}{GitHub repository}.

\subsubsection{Exercise design}
The exercise design is based on the concept of counterfactual simulation. This type of simulation involves presenting users with a queried time series $x_c$, a model's output for that input $\hat{y_c}$, and an explanation of that output $e$. Participants are then asked to predict the model's output $\widetilde{y_c}$ when given a perturbation $p$ of the queried time series $x_c$. We have made a minor change to the concept of counterfactual evaluation to simplify the exercise. In our context, `counterfactual simulation' refers to an experiment designed to quantify whether, through our explanations, a user can predict the trend $t$ of output changes $\widetilde{y_{c}^{t}}$ rather than the specific value of $\widetilde{y_c}$ after a perturbation in a query. 

\begin{figure}[htbp]
\centering
\includegraphics[width=8cm]{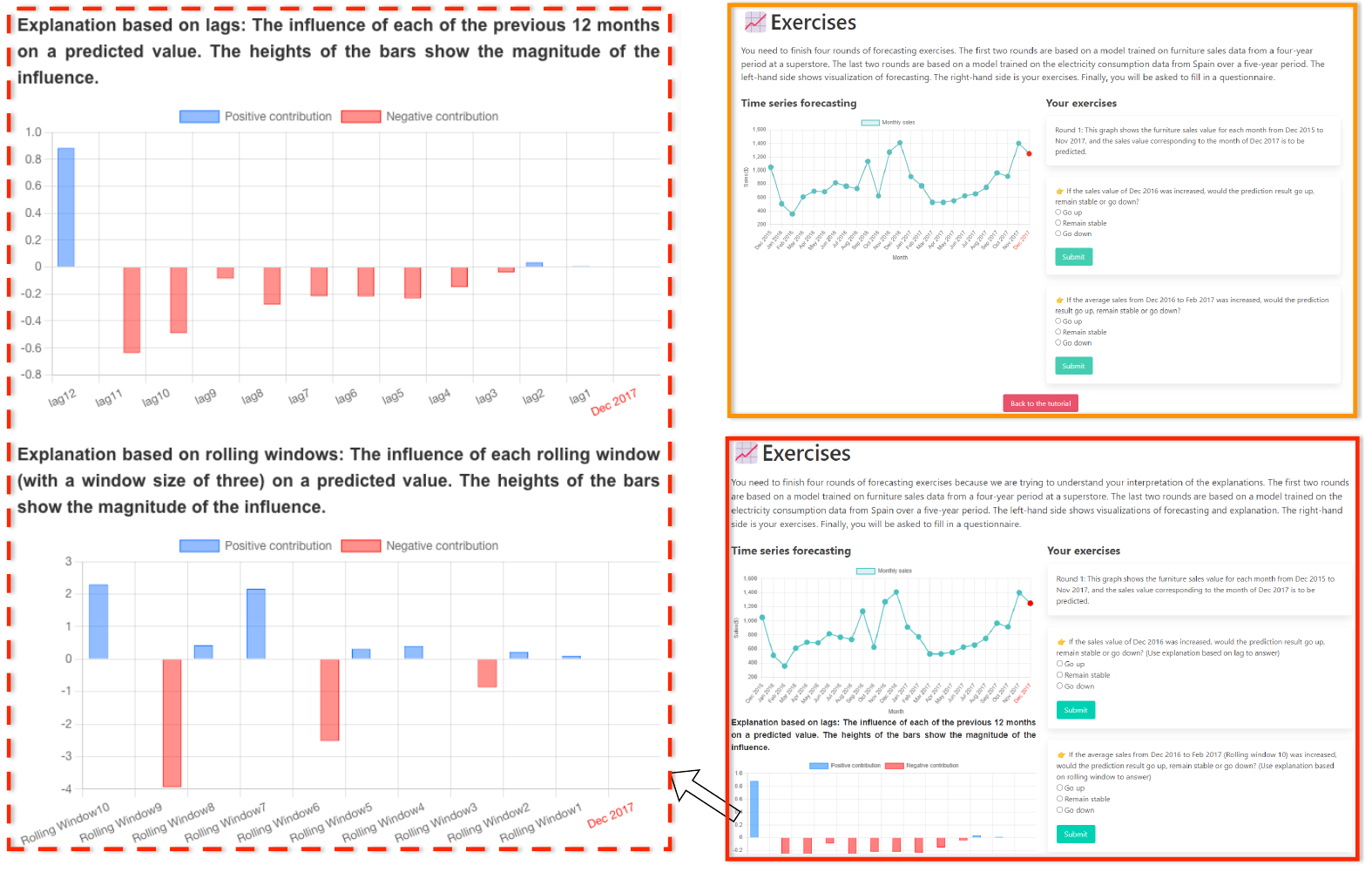}
\caption{Red dotted-line box: zoom in on the explanation for the treatment group; orange solid-line box: control group; red solid-line box: treatment group.}
\label{interface}
\end{figure}

\textbf{Exercise for control group:}
The orange solid-line box in Figure \ref{interface} displays the control group's view. At the top, some basic information is provided. The rest of the page is divided into `Time series forecasting' and `Your exercises'. On the right side, the month marked in red is the one being predicted, based on the data from the previous 12 months. For each exercise, participants are asked to make predictions based on a line chart, without any additional visual explanation. Taking lag as an example, the question is: \textit{``If the sales value of Dec 2016 was increased, would the prediction result go up, remain stable or go down?"} The options are: \textit{Go up, Remain stable, Go down}. After submitting results, participants will proceed to the next round until the `questionnaire' button appears. 

\textbf{Exercise for treatment group:}
The same information and exercise process is provided to participants in the treatment group. The difference is that they are provided with additional visual explanations, as shown in the red dashed-line box on the left side of Figure \ref{interface}. Upon submitting their results, participants will receive a textual explanation of whether their previous answer was incorrect. For instance, the textual explanation might state: \textit{``If a lag has a positive contribution, increasing its value will raise the predicted output by the model, while decreasing it will lower the output. Conversely, if a lag has a negative contribution, increasing its value will lower the predicted output, and decreasing it will raise the output."}.  

\subsubsection{Questionnaire design}
There are 13 questions spanning four categories: basic information, system satisfaction \& the helpfulness of explanations, curiosity \& trust, and overall feedback. These questions are designed based on the criteria for explainable AI proposed by \cite{hoffman2023measures}.
\subsubsection{Participant and procedure}
We evaluated 80 participants per group recruited through the online platform Prolific. Participants were required to be over 18 years old, and fluent in English. Participants from computer science (CS) backgrounds are known to influence user perception of XAI. Therefore, to justify our conclusions and control for relevant variables, we divided each group into subgroups of computer science and non-computer science backgrounds. Each subgroup consisted of 40 participants and was allocated 25 minutes to complete the study. In the control group, participants with computer science backgrounds took an average of 18.08 minutes to complete the tasks (SD = 11.04 minutes). Those without CS background in the same group took slightly less time, averaging 16.79 minutes (SD = 10.84 minutes). In the treatment group, those with a CS background took longer, averaging 19.88 minutes (SD = 8.99 minutes), while those without CS background took the longest time, averaging 21.42 minutes with the highest variability (SD = 13.22 minutes). Each participant was compensated £3.75 (equivalent to £9/hour) for their participation. 
The University of Bristol Faculty of Engineering Research Ethics Committee granted ethics approval for this experiment, under reference code 17907.

Four separate studies for each subgroup were created in Prolific. For each study, participants were prescreened based on their backgrounds. Additionally, to prevent repeat participation, participants were excluded from individual studies in cases where the same participant might access different studies. Once the study was published on Prolific, participants were notified about the studies for which they were eligible, based on the demographic information they had provided. Once a participant has completed the study, their data will be reviewed to decide whether they should be approved and paid, rejected, or returned. 

\subsection{Results}
\subsubsection{Questionnaire evaluation}
\begin{figure}[htbp]
\centering
\subfigure[computer science]{
\begin{minipage}[t]{0.48\linewidth}
\label{q4578controlcs}
\centering
\includegraphics[width=3.5cm]{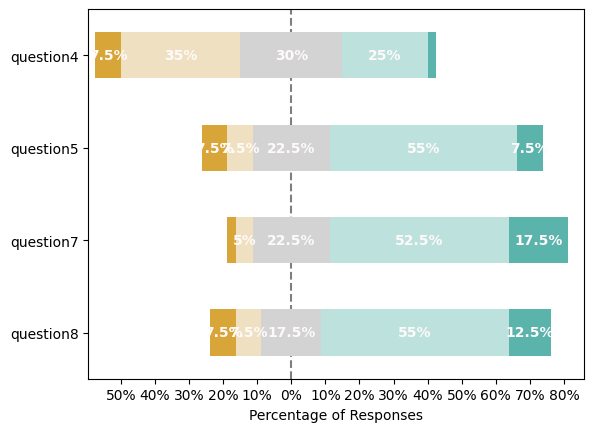}
\end{minipage}%
}%
\subfigure[non-computer science]{
\begin{minipage}[t]{0.48\linewidth}
\label{q4578controlnon}
\centering
\includegraphics[width=3.5cm]{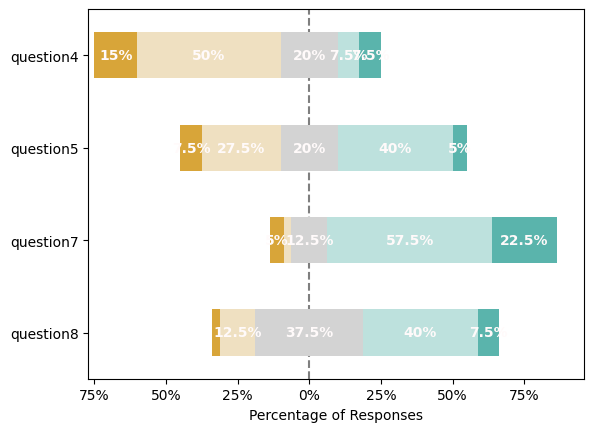}
\end{minipage}%
}%
\centering
\label{q4578control}
\caption{For control group}
\end{figure}

\begin{figure}[htbp]
\centering
\subfigure[computer science]{
\begin{minipage}[t]{0.48\linewidth}
\label{q4578treatcs}
\centering
\includegraphics[width=3.5cm]{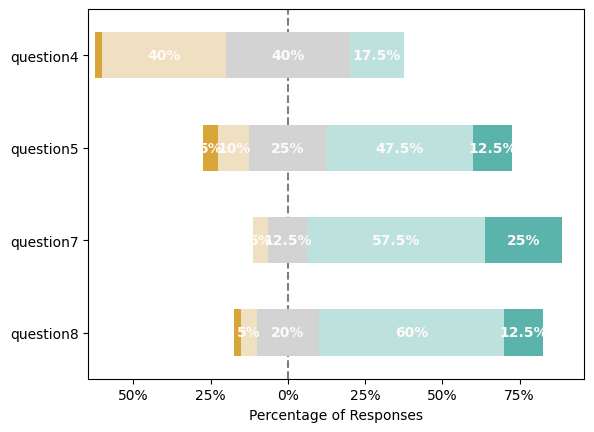}
\end{minipage}%
}%
\subfigure[non-computer science]{
\begin{minipage}[t]{0.48\linewidth}
\label{q4578treatnoncs}
\centering
\includegraphics[width=3.5cm]{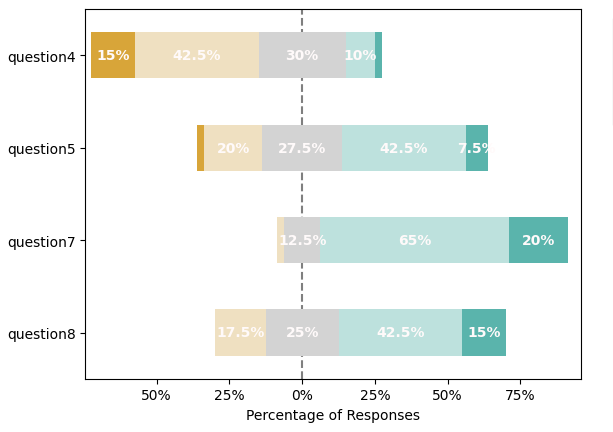}
\end{minipage}%
}%
\centering
\caption{For treatment group}
\end{figure}

Figure \ref{q4578controlcs}, \ref{q4578controlnon}, \ref{q4578treatcs} and \ref{q4578treatnoncs} highlight issues with system satisfaction and explanations helpfulness (Q4, Q5, Q7, Q8) for the control and treatment group, each from different backgrounds, respectively.\footnote{From gold to green is from `very hard' to `very easy' for Q4, `not very helpful' to `very helpful' for Q5,7, `not very satisfied' to `very satisfied' for Q8} 

Question 4 surveyed participants' understanding of lag and rolling window in the field of time series regarding their difficulty. We observed that participants with CS background in both the control group (57.5\%) and the treatment group (57.5\%) found these concepts relatively straightforward, aligning with our expectations. Conversely, more than half of the participants from non-computer science backgrounds found the concepts challenging. Additionally, three times as many non-computer science participants rated the concepts as `very hard' compared to their computer science counterparts. Question 5 assessed how helpful participants found the use of global explanations in understanding time series predictions. Interestingly, a higher percentage of participants with a machine learning background (identified in Question 2) found global explanations to be helpful compared to those without such a background. Global explanations provide an overview of the model’s overall behaviour and decision-making processes. However, those without machine learning background generally seek information that directly impacts their outcomes with the model. Question 6 assessed which explanations were more helpful for the treatment group and whether participants in the control group preferred to have an explanation when making predictions. In the treatment group, about half of the participants, both from computer science (45\%) and non-computer science backgrounds (47.5\%), found both explanations helpful. Among the remaining participants, 25\% from CS background found the `lag' explanation more useful, while 15\% preferred the `rolling window' explanation. Conversely, among those without CS background, 20\% preferred the `lag' explanation, and 22.5\% preferred the `rolling window'. Given these results, we conclude that both explanations are considered helpful regardless of the participants' backgrounds. A definitive conclusion about which explanation is superior cannot be drawn. In the control group, almost all participants requested explanations, except for four in the computer science subgroup. Question 7 focused on system satisfaction across both groups, while Question 8 assessed the overall experience of participants. Figures \ref{q4578controlcs}, \ref{q4578controlnon}, \ref{q4578treatcs} and \ref{q4578treatnoncs} demonstrate that almost all participants found the interface helpful and the exercises meaningful, regardless of their group or background. This indicates that participants were pleased with the experiment and recognized the value of the exercises.

\begin{figure}[htbp] 
\centering
\subfigure[computer science]{
\begin{minipage}[t]{0.48\linewidth}
\label{likert-control-cs}
\centering
\includegraphics[width=3.5cm]{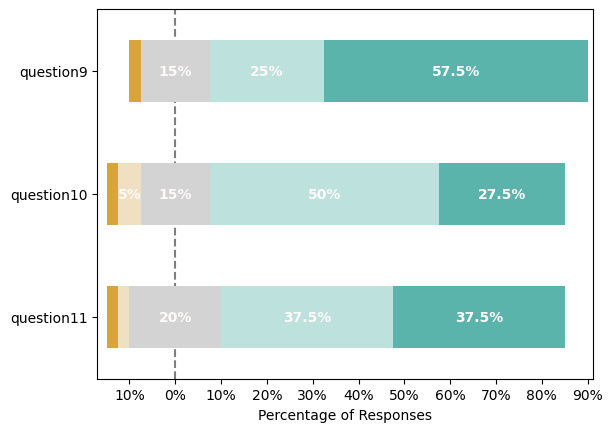}
\end{minipage}%
}%
\subfigure[non-computer science]{
\begin{minipage}[t]{0.48\linewidth}
\label{likert-control-noncs}
\centering
\includegraphics[width=3.5cm]{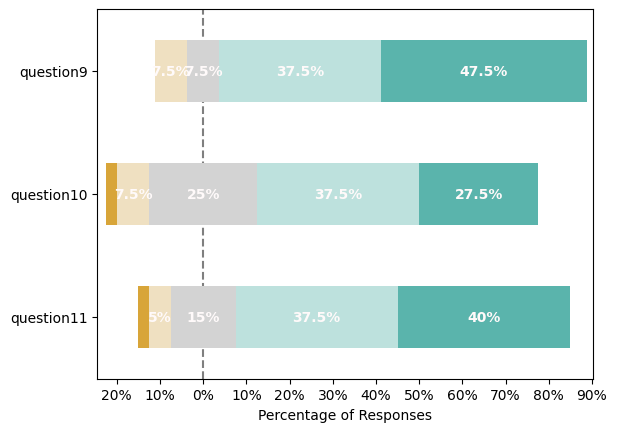}
\end{minipage}%
}%
\centering
\caption{For control group}
\end{figure}

\begin{figure}[htbp] 
\centering
\subfigure[computer science]{
\begin{minipage}[t]{0.48\linewidth}
\label{likert-treatment-cs}
\centering
\includegraphics[width=3.5cm]{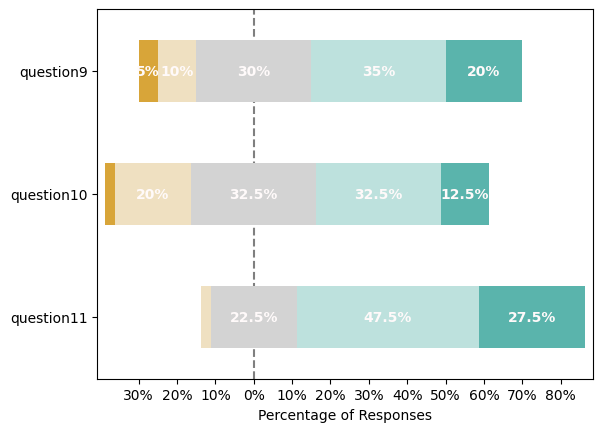}
\end{minipage}%
}%
\subfigure[non-computer science]{
\begin{minipage}[t]{0.48\linewidth}
\label{likert-treatment-noncs}
\centering
\includegraphics[width=3.5cm]{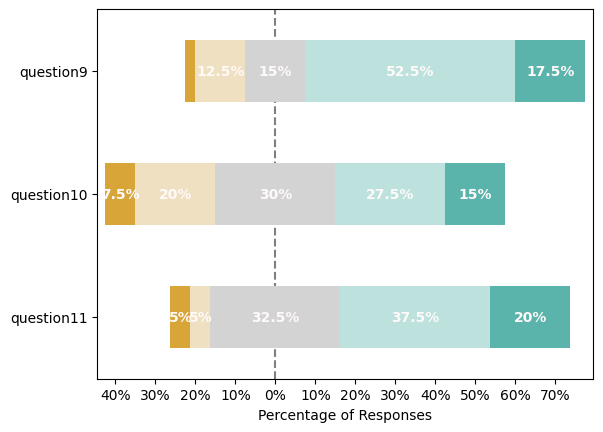}
\end{minipage}%
}%
\centering
\caption{For treatment group}
\end{figure}

Figure \ref{likert-control-cs}, \ref{likert-control-noncs}, \ref{likert-treatment-cs} and \ref{likert-treatment-noncs} highlight issues related to curiosity and trust for both groups, each from different backgrounds, respectively.\footnote{From gold to green is from `disagree strongly' to `agree strongly' for Q9,10,11} For Question 9, which assesses participants' curiosity about understanding the time series forecasting process, only 55\% of participants from the treatment group with CS background expressed curiosity, while 30\% remained neutral. In contrast, over 70\% of participants from other groups expressed a desire to understand the process. This might suggest that participants from computer science focused more on the performance and results of the system rather than on the process itself. Question 10 focused on building confidence through explanations of AI systems. We found that explanations contributed similarly to confidence levels in both computer science and non-computer science participants within the treatment group, with nearly 50\% reporting increased confidence. In contrast, over 50\% of participants from the control group believed that the explanations would enhance their confidence in AI systems. Positive feedback from the treatment group in computer science indicated that `\textit{The visualizations provided regarding the models made the tasks somewhat easier to complete and instilled a bit more confidence in completing them correctly.}' For Question 11, regarding the perceived utility of XAI tools, the control groups, particularly the computer science subgroup, demonstrated a strong belief in the positive impact of XAI.

\subsection{Simulatability results}
In addition to the questionnaire analysis, we also obtained valuable results from exercises, which were analyzed from a statistical perspective. In this study, participants engaged in four rounds of exercises, each round comprising two questions. For each question, participants could earn 1 point for a correct answer and 0 points for an incorrect one, contributing to a total possible score of 2 points per round and 8 points overall for the entire exercise. To determine if there were statistically significant differences in how well the prediction results were across the treatment and control groups, we employed the Mann-Whitney U-Test. This non-parametric test was chosen because it is well-suited for comparing ordinal data across independent groups, particularly when the data are not normally distributed. The \textit{null hypothesis} assumed that there was no difference between the control and treatment groups in terms of total scores achieved, while the \textit{alternative hypothesis} posited a difference. The Mann-Whitney U-Test results showed no significant difference between the control and treatment groups for participants in CS background, with a U value of 794, n1 = 40, n2 = 40, and p = .958, indicating that the null hypothesis could not be rejected. However, among the non-computer science group, a significant difference was observed, with U = 559, n1 = 40, n2 = 40, and p = .021, leading to the rejection of the null hypothesis. These results suggest that the explanations were significantly more effective for participants without CS background.

\section{Conclusion and future work}
\label{conclusionandfuturework}
In summary, we proposed TSFeatLIME, a framework for generating local explanations in univariate time series using auxiliary features and incorporating distance to enhance surrogate model fidelity. We developed a web interface to evaluate explanation effectiveness across different backgrounds, finding them more effective for non-CS participants. While explanations using rolling window and lag techniques were helpful without a clear preference, a designed questionnaire indicated that explanations moderately boost confidence in AI systems.

\bibliographystyle{IEEEtran}
\bibliography{ieee}

\end{document}